\documentclass[sigconf]{acmart}

\usepackage{subcaption}
\usepackage{multirow}
\usepackage{bm}
\usepackage{color}
\def\x{$\times$}
\newlength\savewidth\newcommand\shline{\noalign{\global\savewidth\arrayrulewidth
  \global\arrayrulewidth 1pt}\hline\noalign{\global\arrayrulewidth\savewidth}}
\usepackage{balance}

\newcommand{\RNum}[1]{\uppercase\expandafter{\romannumeral #1\relax}}

\AtBeginDocument{%
  \providecommand\BibTeX{{%
    \normalfont B\kern-0.5em{\scshape i\kern-0.25em b}\kern-0.8em\TeX}}}

\copyrightyear{2021}
\acmYear{2021}
\setcopyright{acmlicensed}\acmConference[MM '21]{Proceedings of the 29th ACM International Conference on Multimedia}{October 20--24, 2021}{Virtual Event, China}
\acmBooktitle{Proceedings of the 29th ACM International Conference on Multimedia (MM '21), October 20--24, 2021, Virtual Event, China}
\acmPrice{15.00}
\acmDOI{10.1145/3474085.3475344}
\acmISBN{978-1-4503-8651-7/21/10}

\begin{document}
\fancyhead{}
\title{DSANet: Dynamic Segment Aggregation Network \\ for Video-Level Representation Learning}

\thanks{$*$ Equal contribution. $\dag$ Corresponding author (wuwenhao17@mails.ucas.edu.cn).}

\author{
    Wenhao Wu$^{1*\dag}$,
    Yuxiang Zhao$^{1,2*}$,
    Yanwu Xu$^{3}$, 
    Xiao Tan$^{1}$,
    Dongliang He$^{1}$, \\
    Zhikang Zou$^{1}$,
    Jin Ye$^{1}$,
    Yingying Li$^{1}$,
    Mingde Yao$^{1}$,
    Zichao Dong$^{1}$,
    Yifeng Shi$^{1}$ 
}

\affiliation{%
    \institution{
    $^1$ Baidu Inc.  \qquad
    $^2$ Shenzhen Institute of Advanced Technology, CAS \qquad
    $^3$ University of Pittsburgh
    }
    \country{}
}


\begin{abstract}
Long-range and short-range temporal modeling are two complementary and crucial aspects of video recognition. 
Most of the state-of-the-arts focus on short-range spatio-temporal modeling and then average multiple snippet-level predictions to yield the final video-level prediction. 
Thus, their video-level prediction does not consider spatio-temporal features of how video evolves along the temporal dimension.
In this paper, we introduce a novel \emph{\textbf{D}ynamic \textbf{S}egment \textbf{A}ggregation} (\textbf{DSA}) module to capture relationship among snippets.  
To be more specific, we attempt to generate a dynamic kernel for a convolutional operation to aggregate long-range temporal information among adjacent snippets adaptively. 
The DSA module is an efficient plug-and-play module and can be combined with the off-the-shelf clip-based models (\emph{i.e.,} TSM, I3D) to perform powerful long-range modeling with minimal overhead.
The final video architecture, coined as DSANet.
We conduct extensive experiments on several video recognition benchmarks (\emph{i.e.}, Mini-Kinetics-200, Kinetics-400, Something-Something V1 and ActivityNet) to show its superiority. 
Our proposed DSA module is shown to benefit various video recognition models significantly. For example, equipped with DSA modules, the top-1 accuracy of I3D ResNet-50 is improved from 74.9\% to 78.2\% on Kinetics-400.
Codes are available at \url{https://github.com/whwu95/DSANet}.
\end{abstract}




\maketitle

\section{Introduction}

With the rapid development of the Internet and mobile devices, video data has exploded over the past years. Video action recognition, as a fundamental problem in video analytics, has become one of the most active research topics. 
\begin{figure}[t]
    \centering
    \includegraphics[width=0.48\textwidth]{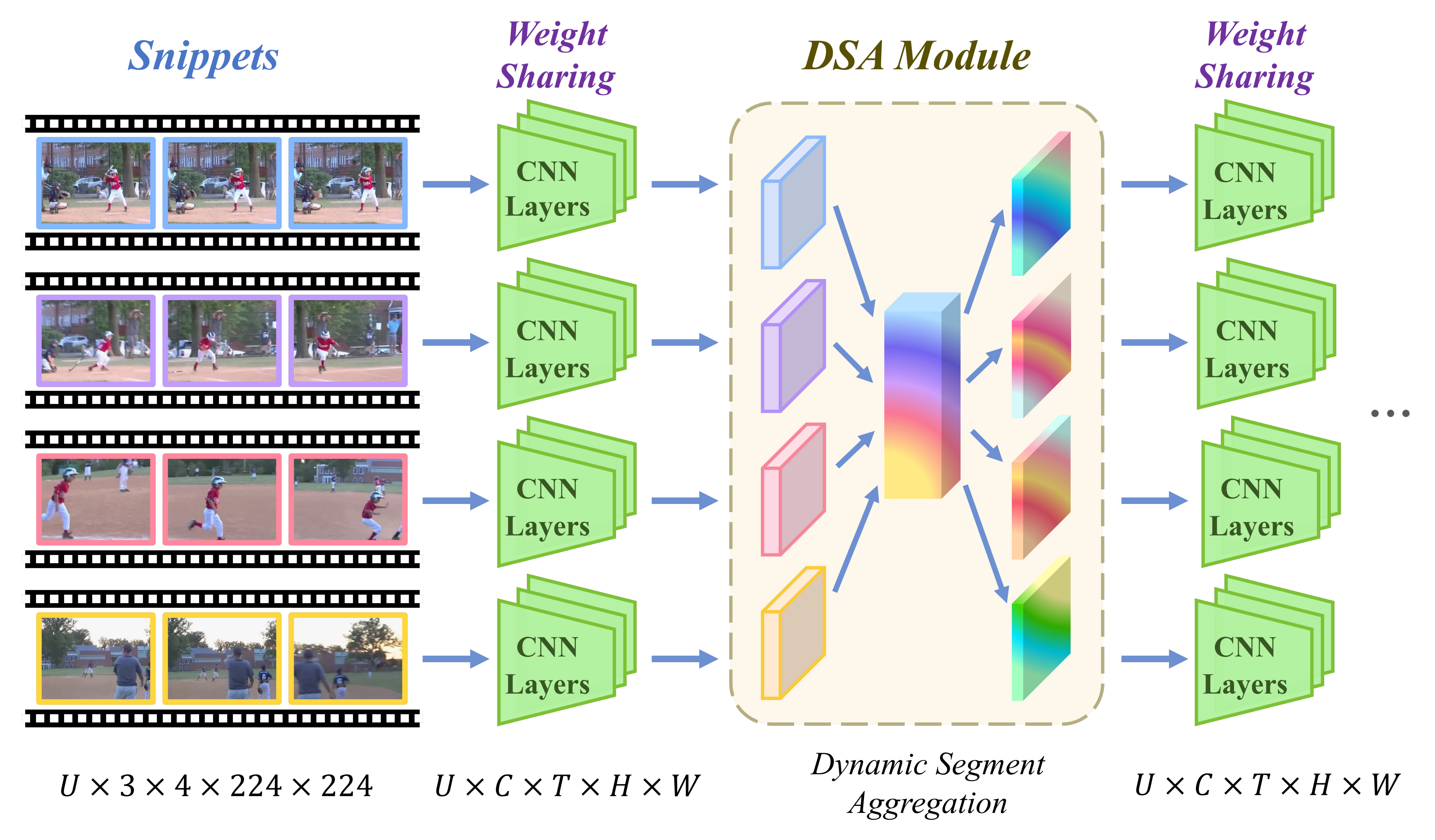}
    \caption{
    The pipeline of our method for video recognition. Input video is divided into several segments and a short snippet is randomly sampled from each segment. Green trapezoids denote the snippet-level spatio-temporal networks (\emph{e.g.}, I3D, TSM) which share parameters on all snippets. DSA module can be easily inserted into the networks to capture long-range temporal structure.
    }
    \label{fig:overview}
\end{figure} 
Previous methods have obtained noticeable improvements for video recognition.
3D CNNs are intuitive spatio-temporal networks and natural extension from their 2D counterparts, which directly tackle 3D volumetric video data~\cite{c3d,i3d}. However, the heavy computational cost of 3D CNNs limits its application. Recent studies have shown that factorizing 3D convolutional kernel into spatial (\emph{e.g.}, 1\x3\x3) and temporal components (\emph{e.g.}, 3\x1\x1) is preferable to reduce complexity as well as boost accuracy, \emph{e.g.}, P3D~\cite{p3d}, R(2+1)D~\cite{r2+1d}, S3D-G~\cite{s3d}, \emph{etc}. 
In practice, however, compared with their 2D counterparts, the increased computational cost cannot be ignored.
Thus, 3D channel-wise convolution are also applied to video recognition in CSN~\cite{CSN} and X3D~\cite{feichtenhofer2020x3d}.
To capture temporal information with reasonable training resources, the other alternative 2D CNN based architectures are developed \emph{i.e.,} TSM~\cite{tsm}, TEI~\cite{teinet}, TEA~\cite{li2020tea}, MVFNet~\cite{wu2020MVFNet}, \emph{etc}. There methods aim to design efficient temporal module based on existing 2D CNNs to perform efficient temporal modeling.
Temporal Shift Module (TSM)~\cite{tsm} is the representative 2D-based model, which achieves a good trade-off between performance and complexity meanwhile introduces zero FLOPs for the 2D CNN backbone.

However, the above architectures, both based on 2D CNNs and 3D CNNs, are mainly designed to learn short-range spatio-temporal representation (snippet-level) instead of long-range representation (video-level).
Specifically, during training, these snippet-level methods sample sub-frames (4 frames, 8 frames or 16 frames) from a random snippet of the original video.
While for inference, they uniformly sample multiple short snippets from a video and compute the classification scores for all snippets individually and the final prediction is just the average voting from the snippet-level predictions.
Thus, during training, these clip-based models take no account of modeling long-range temporal structure due to the limited observation of only a single snippet.
Also, simply averaging the prediction scores of all snippets could yield a sub-optimal solution during inference.

To alleviate the problem mentioned above, Temporal Segment Network~\cite{tsn} provides a simple framework to perform video-level learning, which just simply fuses snippet-level predictions. Lately, V4D~\cite{zhang2020v4d} presents an intuitive 4D convolutional operation to capture inter-clip interaction, which is a natural extension of enhancing the representation of the original clip-based 3D CNNs. However, TSN fails on capturing the finer temporal structure, (\emph{e.g.}, the temporal pattern of the ordering of features), and V4D causes high computational cost compared with the TSN-equipped counterpart and might lack the ability to describe the dynamics by simply employing a fixed number of video invariant kernels. This inspires us to focus on designing a temporal module, which keeps both advantages of high efficiency and strong ability of capturing long-range temporal structure.

In this paper, we creatively introduce a module to adaptively aggregate the features of multiple snippets, namely \emph{\textbf{D}ynamic \textbf{S}egment \textbf{A}ggregation Module} (\textbf{DSA} Module). 
Figure~\ref{fig:overview} illustrates the pipeline of our solution, DSA module is a novel generic module that can be easily plugged into the existing clip-based networks (\emph{e.g.,} TSM~\cite{tsm}, I3D~\cite{i3d}) to perform long-range modeling with the ignorable cost. 
Briefly, the DSA module firstly gets global context information from features of all snippets, then it utilizes the contextual view to learns the weight of aggregation in a channel-wise convolution manner.
Thus, DSA module can capture relation-ship among snippets via dynamic fusion along snippet dimension.
To make this module more efficient, we separate the features across the channel dimension. One keeps the original activation and the other performs dynamic segment aggregation separately followed by a concatenation operation.
In this way, DSA only introduces a bit more FLOPs which is close to zero, while bringing remarkable improvement. 
In practice, we integrate the DSA module into the existing spatio-temporal block (\emph{e.g.}, TSM block, I3D block) to form our DSANet.

We identify the effectiveness of the proposed DSA module via comprehensive ablations. Extensive experimental results on multiple well-known datasets, including Mini-Kinetics-200~\cite{s3d}, Kinetics-400 \cite{kay2017kinetics}, Something-Something V1 \cite{sth-sth} and ActivityNet-v1.3 \cite{caba2015activitynet} show the superiority of our solution. All of these prove DSA boosts the performance over its snippet-level counterpart.
Overall, our major contributions are summarized as follows:

\begin{itemize}
    \item Instead of snippet-level temporal modeling, we propose to exploit an effective and efficient video-level framework for learning video representation. To tackle this, the proposed DSA module provides a novel mechanism to adaptively aggregate snippet-level features.
    \item The DSA module works in a plug-and-play way and can be easily integrated into existing snippet-level methods. Without any bells and whistles, the DSA module brings consistent improvements when combined with both 2D CNN-based and 3D CNN-based networks (\emph{e.g.}, TSM, I3D, \emph{etc}).
    \item Extensive experiments on four public benchmark datasets demonstrate that the proposed DSA obtain an evident improvement over previous long-range temporal modeling methods with a minimal computational cost.
\end{itemize}

\section{Related Work}
\subsection{CNNs for Video Recognition.}
Using cutting edge 2D CNN architecture is desirable to address the video recognition task. 
Two-stream\cite{two-stream} proposes to jointly learn spatial and temporal context through two parallel branch structure. 
3D CNN expands a new dimension of 2D convolution whereby intuitively apply to assemble spatio-temporal feature. C3D\cite{c3d} is the first one to using VGG of 3D type in video recognition. 
I3D \cite{i3d} is a milestone of 3D convolution which used pre-trained parameters of 2D CNNs onto the 3D CNNs by inflating the parameter on a new axis, boost the performance by a large margin. 
Along with significant study on separable convolution, some methods (\emph{e.g.,} S3D-G~\cite{s3d}, R(2+1)D~\cite{r2+1d}, P3D \cite{p3d}) decompose the full 3D convolution into a 2D convolution and a 1D temporal convolution to reduce the burdensome parameters and accelerate the training. 

To alleviate the high computational cost of video classification networks, the other alternative 2D CNN based architectures (\emph{e.g.,} STM~\cite{stm}, TSM~\cite{tsm}, ECO~\cite{eco}, TEI~\cite{teinet}, TEA~\cite{li2020tea}, GSM~\cite{GSM}, MVFNet~\cite{wu2020MVFNet}) and channel-wise 3D CNN based architectures (\emph{e.g.,} CSN~\cite{CSN}, X3D~\cite{feichtenhofer2020x3d}) are developed. 
Also, there is active research on dynamic inference~\cite{wu2020dynamic,wang2021adaptive}. These methods can get good computation/accuracy trade-offs. 

Besides, recently we notice that the concurrent work to ours is TANet~\cite{liu2020tam}, which composed of two branches, the local branch aims to learn a location sensitive importance map to enhance discriminative features and the global branch produces the location invariant weights to aggregate temporal information. Both the global branch of TANet and our DSANet are inspired by the previous dynamic convolution mechanisms~\cite{chen2020dynamic,wang2019carafe}. However, the TANet aims to capture \emph{clip-level} temporal information and our DSANet targets on \emph{video-level} temporal information.

\subsection{Long-Term Modeling for Video Recognition}
The obvious difference between long-term modeling method and short-term modeling method is whether the method can capture the relation of different video segments for video-level representation learning.
Various long-term modeling frameworks have been developed for capturing more complex temporal structure in video action recognition fields.
In ~\cite{LRCN}, CNN features of video frames are fed into the LSTM network for video recognition.
ShuttleNet~\cite{shuttleNet} introduces biologically-inspired feedback connections to model long-term dependencies of spatial CNN descriptors. 
NetVlad~\cite{netvlad}, ActionVlad~\cite{actionvlad}, AttentionClusters~\cite{attentioncluster} are proposed for better local feature integration instead of directly average pooling as used. 
MARL\cite{wu2019multi} uses multiple agents as frame selectors instead of the general uniform sampler from the entire video for better global temporal modeling. 
Each agent learns a flexibly moving policy through the temporal axis to get a vital representation frame and other agents' behavior as well.
\cite{korbar2019scsampler} proposes a new solution as dynamic sampler. Ranking the high score clips generated by a small classifier network, a top-k pooling method is then applied to feed the most related parts into a large network. 

\begin{figure*}[t]
    \centering
    \includegraphics[width=1\textwidth]{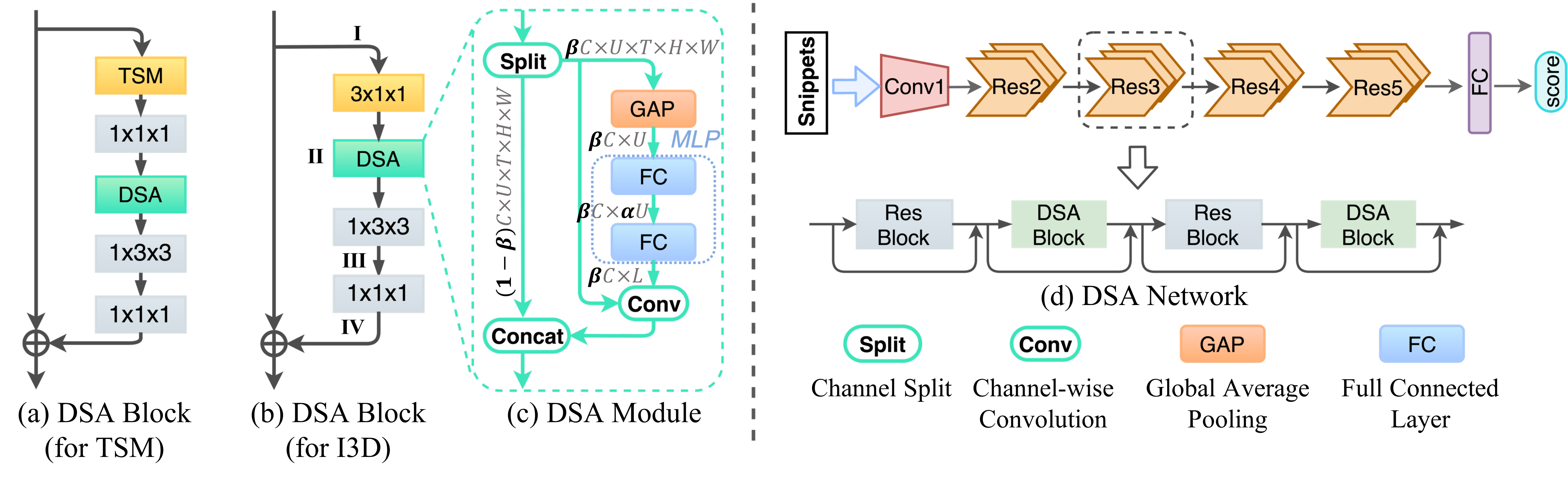}
    \caption{Illustration of our framework for video action recognition. (a) shows the TSM block equipped with a DSA module. (b) shows the I3D block equipped with a DSA module. (c) depicts the architecture of the DSA module. (d) shows our DSA block which integrates the DSA module into the ResNet-Style block.}
    \label{fig:dsa}
\end{figure*} 

Closely related works are TSN~\cite{tsn}, T-C3D\cite{tc3d}, StNet\cite{he2019stnet} and V4D~\cite{zhang2020v4d}.
\cite{tsn} operates on a sequence of short snippets sparsely sampled from the entire video and the parameter is updated according to segmental consensus derived from all snippet-level prediction. T-C3D replaces 2D backbones with C3D. So these models only capture coarse co-occurrence rather than the finer temporal structures. StNet applies 2D ConvBlocks on ``superimage'' for local spatial-temporal modeling and 1D temporal convolutions for global temporal dependency. V4D~\cite{zhang2020v4d} tends to focus on the intuitive 4D convolutional operation to capture inter-clip interaction, which is a natural extension to enhance the representation power of the original snippet-level 3D CNNs, however, causes high computational cost.
Therefore we present a flexible long-range modeling module that shares the capacity of learning video-level representations yet still keeps the efficiency of the snippet-level counterpart.


\section{Dynamic Segment Aggregation}

\subsection{Preliminary: 4D Convolution}
To aggregate the snippet-level features, V4D cast 3D convolution into 4D convolution. Formally, following the notations in V4D~\cite{zhang2020v4d}, the whole video formed by a series of short snippets of $A_{i} \in \mathbb{R}^{C \times T \times H \times W}$, where $C$ is the number of the input channel, $C'$ denotes the number of output channel, $T$ is the dimension of sampled frames from each snippet, and $H$, $W$ represent height and width of the frames respectively.
Further, video can then be represented as a 5-D tensor $V:=\left\{A_{1}, A_{2}, \ldots, A_{U}\right\}$, where $U$ is the number of snippets as a new dimension. Then, the 4D convolutional operation can be formulated as:

\begin{equation}
O_{c^{\prime}, u, t, h, w}=\sum_{c, i, j, k,l} K_{c^{\prime}, c, l, k, i, j} \   V_{c, u+\hat{l}, t+\hat{k}, h+\hat{i}, w+\hat{j}},
\end{equation}
where $O$ is the output tensor, $K$ is the 4D convolution kernel, $i, \ j, \ k, \ l$ index along the $H$, $W$, $T$, $U$ dimensions of kernel and $c'$ represents the number of kernel.

\subsection{Dynamic Kernel Generation}
Due to the flowing dynamics of videos, we learn an adaptive, context-aware kernel to handle video variations along the temporal dimension.
To be simplified, we denote the input to the DSA module as a tensor $V \in \mathbb{R}^{ C \times U \times T \times H \times W}$.
The overall DSA module is illustrated in Figure 2(c).
We can formalize this dynamic kernel generation as follows:

\begin{equation}
    K = Softmax(\mathcal{F}(\theta, g(V))),
\end{equation}
where $g$ aggregates video feature maps $V$ for global view by using global average pooling, which produces ${\hat{V}} \in \mathbb{R}^{ C \times U}$.
$\mathcal{F}$ is a mapping function that could generate kernel $K$ with learnable parameters of $\theta$.
And $K \in \mathbb{R}^{C \times L} $ is generated dynamic convolutional kernel for each channel, where $L$ is the kernel size. In this paper, we set $L$ to 3.
To capture context information effectively, we use multi-layer perceptron (MLP) network with batch normalization and activation function (\emph{e.g.}, ReLU) to instantiate $\mathcal{F}$ in this paper.
$\alpha$ is a super-parameters of fully connected layer in MLP to control the size of feature.
To conduct better capability, we constraint $K$ with a \emph{Softmax} function to generate normalized kernel weights.

	\newcommand{\blocks}[3]{\multirow{3}{*}{\(\left[\begin{array}{c}\text{1$\times$1$^\text{2}$, #2}\\[-.1em] \text{1$\times$3$^\text{2}$, #2}\\[-.1em] \text{1$\times$1$^\text{2}$, #1}\end{array}\right]\)$\times$#3}
	}
	\newcommand{\blockt}[3]{\multirow{3}{*}{\(\left[\begin{array}{c}\text{\underline{3$\times$1$^\text{2}$}, #2}\\[-.1em] \text{1$\times$3$^\text{2}$, #2}\\[-.1em] \text{1$\times$1$^\text{2}$, #1}\end{array}\right]\)$\times$#3}
	}
	
	\newcommand{\blocksb}[2]{\multirow{3}{*}{\(\left[\begin{array}{c}\text{1$\times$3$^\text{2}$, #1}\\[-.1em] \text{1$\times$3$^\text{2}$, #1}\end{array}\right]\)$\times$#2}
	}
	
	\newcommand{\blocktb}[2]{\multirow{3}{*}{\(\left[\begin{array}{c}\text{3$\times$3$^\text{2}$, #1}\\[-.1em] \text{3$\times$3$^\text{2}$, #1}\end{array}\right]\)$\times$#2}	
	}

	\begin{table}[t]
		\centering
		\caption{\textbf{Two backbone instantiations of the I3D network}. 
			The dimensions of kernels are denoted by $\{$$T$\x $S^2$, $C$$\}$ for temporal, spatial, and channel sizes.
			Strides are denoted as $\{$temporal stride, spatial stride$^2$$\}$.
		}		
		\scalebox{1.0}{
		\setlength{\tabcolsep}{2.0pt}
			\begin{tabular}{c|c|c|c}
				stage & I3D ResNet18 &  I3D ResNet50 & output sizes\\
				\shline
				\multirow{1}{*}{raw clip} & - & - & 32\x224$^\text{2}$ \\
				\hline
				\multirow{2}{*}{data layer} & \multirow{2}{*}{stride 8, 1$^\text{2}$} & \multirow{2}{*}{stride 8, 1$^\text{2}$} &  \multirow{2}{*}{4\x$224^2$}   \\
				&  &  \\
				\hline
				\multirow{2}{*}{conv$_1$} & \multicolumn{1}{c|}{1\x7$^\text{2}$, {64}} & \multicolumn{1}{c|}{1\x7$^\text{2}$, {64}} &  \multirow{2}{*}{4\x$112^2$}    \\
				& stride 1, 2$^\text{2}$ & stride 1, 2$^\text{2}$  \\
				\hline
				\multirow{2}{*}{pool$_1$}  & \multicolumn{1}{c|}{1\x3$^\text{2}$ max} & \multicolumn{1}{c|}{1\x3$^\text{2}$ max} &  \multirow{2}{*}{4\x$56^2$}  \\
				& stride 1, 2$^\text{2}$ & stride 1, 2$^\text{2}$ & \\
				\hline
				\multirow{3}{*}{res$_2$} & \blocksb{{64}}{2} & \blocks{{256}}{{64}}{3} & \multirow{2}{*}{4\x$56^2$}   \\
				&  & \\
				&  & \\
				\hline
				\multirow{3}{*}{res$_3$} & \blocksb{{128}}{2} &  \blocks{{512}}{{128}}{4}  & \multirow{2}{*}{4\x$28^2$}   \\
				&  & \\
				&  & \\
				\hline
				\multirow{3}{*}{res$_4$} & \blocktb{{256}}{2} & \blockt{{1024}}{{256}}{6} &  \multirow{2}{*}{4\x$14^2$}   \\
				&  & \\
				&  & \\
				\hline
				\multirow{3}{*}{res$_5$} & \blocktb{{512}}{2} & \blockt{{2048}}{{512}}{3} &   \multirow{2}{*}{4\x$7^2$}  \\
				&  & \\
				&  & \\
				\hline
				\multicolumn{3}{c|}{global average pool, fc}  & \# classes \\
		\end{tabular}
		}
		\label{tab:arch}
	\end{table}

\subsection{DSA as a Specific Case of 4D Convolution}
In this section, we formulate our proposed DSA as a specific case of 4D convolution.
While capturing the long-term perception structure, the learned kernel $K$ above can be viewed as aggregation weights to perform dynamic weighted fusion along $U$ dimensions via channel-wise convolutional operation

\begin{equation}
    O_{c,u,t,h,w}=\sum_{l}K_{c,l}V_{c,u+l,t,h,w}.
\end{equation}

At this point, we can see that DSA has a reduced kernel size along $T,H,W$ dimensions, which is also a 4D convolutional kernel.

  To make this module more efficient, we decompose the feature maps into two parts as shown in Figure 2(c). Let $V_{1}\in\mathbb{R}^{\beta C \times U \times T \times H \times W}$ and $V_{2} \in \mathbb{R}^{ (1-\beta)C \times U \times T \times H \times W}$ be the two splits of $V$ along the channel dimension, where $\beta \in [0, 1]$ denoted the proportion of input channels for dynamic aggregation. Finally, we get the DSA module output $Y \in \mathbb{R}^{ C \times U \times T \times H \times W}$ as:
\begin{equation}
    Y_{c,u,t,h,w}=\mathit{Concat}(\sum_{l} K_{c,l} V_{1_{c,u+l,t,h,w}},
    V_{2}),
\end{equation}
where \emph{Concat} represents concatenation operation along dimension of channel.

\subsection{DSA Network Architecture}
The proposed DSA module is a flexible module that can be easily plugged into the existing clip-based networks to capture long-range temporal structure. 
Here we illustrate how to adapt it into a residual style block, and then build up the \emph{Dynamic Segment Aggregation Network} (DSANet).
To evaluate the generality of the DSA module, we build up the DSANet on both 2D and 3D backbone networks (\emph{i.e.}, TSM, I3D). 
TSM~\cite{tsm} and I3D~\cite{slowfast}, the two widely used networks, are respectively the representatives of the 2D CNN-based model and 3D CNN-based model.
To avoid confusion, as shown in Table~\ref{tab:arch}, I3D refer to the slow-only branch of SlowFast~\cite{slowfast} which has no temporal downsampling operations. 

For a fair comparison with many recent state-of-the-art video recognition methods, we follow the same practices and instantiate the DSANet using ResNet-style backbone in this instantiation. 
Specifically, we integrate the proposed module into the I3D ResNet block or TSM ResNet block to construct our DSA block. The overall design of the DSA block is shown in Figure 2(a) and Figure 2(b). And there are four optional positions (denoted as \RNum{1}, \RNum{2}, \RNum{3}, \RNum{4}) to insert DSA module and we will perform ablation studies in Sec.~\ref{sec:ab}.
Then following Figure 2(d), we replace some residual blocks with DSA blocks to form our DSANet.


\subsection{Discussion}
We have noticed that our DSA module gets global context information from features of all snippets by \emph{global average pooling}, which is same with the \emph{squeeze operation} in SENet~\cite{senet}.
However, there are essential differences between SENet and DSA as follows:

1) We propose to aggregate the snippet-level features by utilizing an efficient \textbf{convolutional operation}.

2) In corresponding to the dynamics of video data, we further build a method to learn an adaptive convolutional kernel with ignorable computation cost for the convolutional operation. Essentially, our method is different from SENet, which applies a \textbf{self-attention} mechanism to generate a channel-wise attention.

\section{Experiments}
\subsection{Datasets and Evaluation Metrics}
We evaluate our method on several standard video recognition benchmarks, including Mini-Kinetics-200~\cite{s3d}, Kinetics-400~\cite{kay2017kinetics}, Something-Something V1~\cite{sth-sth} and ActivityNet-v1.3~\cite{caba2015activitynet}. 
Kinetics-400 contains 400 human action categories and provides around 240k training video clips and 20k validation video clips. Each example is temporally trimmed to be around 10 seconds.
Mini-Kinetics-200 consists of the 200 categories with the most training examples and is a subset of the Kinetics-400 dataset. Since the full Kinetics-400 dataset is quite large, we use Mini-Kinetics-200 for our ablation experiments.
For Something-Something V1 datasets, the actions therein mainly include object-object and human-object interactions, which require strong temporal relation to well categorizing them. Something-Something V1 includes about 110k videos for 174 fine-grained classes.
To elaborately study the effectiveness and generalization ability of our solution, we also conduct experiments for untrimmed video recognition. We choose ActivityNet-v1.3 which is a large-scale untrimmed video benchmark, contains 19,994 untrimmed videos of 5 to 10 minutes from 200 activity categories.

We report top-1 and top-5 accuracy (\%) for Mini-Kinetics-200, Kinetics-400 and Something-Something V1. For ActivityNet-v1.3, we follow the original official evaluation scheme using mean average precision (mAP).
Also, we report the computational cost (in FLOPs) as well as the number of model parameters to depict model complexity. In this paper, we \textbf{only} use the RGB frames of these datasets and don't use extra modalities (\emph{e.g.,} optical flow, audio).

\subsection{Implementation Details}
Here we report the implementation details.

\begin{table*}
		\caption{{Ablation studies} on \textbf{Mini-Kinetics-200}. We show top-1 and top-5 classification accuracy (\%), as well as computational complexity measured in FLOPs (floating-point operations).
		}
		\begin{subtable}[th]{0.72\textwidth}
		\centering
		\caption{\textbf{Study on the effectiveness of DSA module}. $T$ denotes the number of frames sampled from each video snippet, $U$ denotes the number of snippets. Backbone: I3D R18.}		
			\begin{tabular}{c|c|c|ccc}
			\shline
				\multicolumn{1}{c|}{Model} & $T_{train}$\x$U_{train}$ & $T_{infer}$\x$ U_{infer}$\x\#crop & Top-1 & Top-5 & Params  \\
				\hline
				I3D R18 & {$4 \times  1$} & {$4 \times 10 \times 3$} & 72.2  & 91.2 & 32.3M   \\
				I3D R18& {$16 \times  1$} & {$16 \times 10 \times 3$} & 73.4 & 91.1 & 32.3M  \\
				TSN+I3D R18& {$4 \times 4$} & {$4 \times 10 \times 3$} & 73.0 & 91.3 & 32.3M \\
				V4D+I3D R18 & {$4 \times 4$} & {$4 \times 10 \times 3$} & 75.6 & 92.7 & 33.1M \\
				\hline
				DSA+I3D R18 & {$4 \times 4$} & {$4 \times 8 \times 3$} & \textbf{77.3} & \textbf{93.9} & 32.3M  \\ \shline
			\end{tabular}   
			\label{tab:ablation:r18_perf}
		\end{subtable}	
	    \hspace{4mm}
		\begin{subtable}[th]{0.24\textwidth}
		\centering
		\caption{\textbf{Study on different position to insert DSA module}. Setting: I3D R50, $\alpha$=2, $\beta$=1, stage: res$_5$. }		
			\begin{tabular}{c|cc}
			\shline
			Position  & Top-1       & Top-5     \\ \hline
				\RNum{1}  & 81.4 & \textbf{95.4}  \\
				\RNum{2}  & \textbf{81.5} & 95.2  \\
				\RNum{3}  & 80.8 & 95.2  \\
				\RNum{4}  & 81.4 & 95.1   \\
				\shline
			\end{tabular}
			\label{tab:ablation:position}
		\end{subtable}
		\\[7pt]
		\begin{subtable}[th]{0.22\textwidth}
		\centering
		\caption{\textbf{Parameter choices of $\bm{\alpha}$}. Setting: I3D R50, Position \RNum{2}, $\beta$=1, inserted stage: res$_5$.}		
			\begin{tabular}{l|cc}
			\shline
			Setting   & Top-1       & Top-5     \\ \hline
				$\alpha$=1  & 81.0 & 95.1  \\
				$\alpha$=2  & \textbf{81.5} & \textbf{95.2}   \\
				$\alpha$=4  & 81.2 & 95.0  \\
				$\alpha$=8  & 81.3 & 95.0  \\ \shline
			\end{tabular}
			\label{tab:ablation:alpha}
		\end{subtable}
		\hspace{2mm}
		\begin{subtable}[th]{0.23\textwidth}
		\centering
		\caption{\textbf{The DSA blocks in different stage of I3D R50}. Setting: Position \RNum{2}, $\alpha$=2, $\beta$=1.}		
			\begin{tabular}{l|cc}
			\shline
			Stage   & Top-1       & Top-5     \\ \hline
				res\{2\}  & 81.4 & 94.7  \\
				res\{3\}  & 81.3 & 95.1   \\
				res\{4\}  & 81.3 & \textbf{95.3}  \\
				res\{5\}  & \textbf{81.5} & 95.2  \\ \shline
			\end{tabular}
			\label{tab:ablation:stage}
		\end{subtable}
		\hspace{2mm}
		\begin{subtable}[th]{0.23\textwidth}
		\centering
		\caption{\textbf{Parameter choices of $\bm{\beta}$}. Setting: I3D R50, Position \RNum{2}, $\alpha$=2, inserted stage: res$_{5}$.}		
			\begin{tabular}{l|cc}
			\shline
			Setting   & Top-1       & Top-5     \\ \hline
				$\beta$=1  & 81.5 & 95.2  \\
				$\beta$=1/2  & \textbf{81.7} & \textbf{95.4}  \\
				$\beta$=1/4  & 81.6 & 95.0  \\
				$\beta$=1/8  & 81.5 & 95.0 \\ \shline
			\end{tabular}
			\label{tab:ablation:beta}
		\end{subtable}		
		\hspace{2mm}
		\begin{subtable}[th]{0.24\textwidth}
		\centering
			\caption{\textbf{The number of DSA block inserted into I3D R50}. Setting: Position \RNum{2}, $\alpha$=2, $\beta$=1/8.}		
		\setlength{\tabcolsep}{2.0pt}
			\begin{tabular}{l|c|cc}
			\shline
			Stages & Blocks  & Top-1       & Top-5     \\ \hline
				res\{5\} & 1 & 81.5 & 95.0  \\
				res\{4,5\} & 4 & 81.5 & 95.3   \\
				res\{3,4\} & 5 & \textbf{81.8} & \textbf{95.4}  \\
				res\{2,3\} & 3 & 81.4 & 95.1  \\ \shline
			\end{tabular}
			\label{tab:ablation:multi-stage}
		\end{subtable}	
		\\[7pt]
		\begin{subtable}[th]{0.29\textwidth}
		\centering
			\caption{\textbf{Different short-term temporal structure for DSA module}.}		
			\begin{tabular}{c|cc}
			\shline
			Model   & Top-1       & Top-5     \\ \hline
			    TSM R50 &  77.4 & 93.4  \\
			    DSA+TSM R50  & \textbf{80.4} & \textbf{95.0}   \\  \hline
			    I3D R50 &  78.0 & 93.9  \\
				DSA+I3D R50  & \textbf{81.8} & \textbf{95.4}  \\ \shline
			\end{tabular}
			\label{tab:ablation:short-term}
		\end{subtable}			
		\hspace{3mm}
	    \begin{subtable}[th]{0.33\textwidth}
	    \centering
            \caption{\textbf{Study on the effectiveness of DSA module with different backbones (I3D R18, I3D R50)}. SENet+I3D uses SE module to replace the DSA module in DSANet.}	    
	    	\begin{tabular}{c|ccc}
            \shline
            Arch.  & I3D & SENet+I3D & DSA+I3D\\
            \hline
            {ResNet18} &  72.2 & 73.8 & \textbf{77.3} \\
            {ResNet50} & 78.0 & 78.5 & \textbf{81.8} \\
            \shline
            \end{tabular}
            \label{tab:ablation:backbone}
	    \end{subtable}
	    \hspace{3mm}
	    \begin{subtable}[h]{0.32\textwidth}
	    \centering
        \caption{\textbf{Training FLOPs}. Comparison with V4D, the extra computation cost brought by the DSA module is close to zero.}	    
	    \setlength{\tabcolsep}{2.0pt}
	    	\begin{tabular}{c|c|cc}
            \shline
            Model  &   Input size & FLOPs\\
            \hline
            {TSN+I3D R50} &  {$4\times4\times224^2\times3$} & 83.8G\\
            {V4D+I3D R50} &  {$4\times4\times224^2\times3$}  & 143.0G\\ \hline
            {DSA+I3D R50} &  {$4\times4\times224^2\times3$} &  83.8G \\
            \shline
            \end{tabular}
            \label{tab:ablation:flops}
	    \end{subtable}		
		\label{tab:ablations}
	\end{table*}
\textbf{Training.} 
We evaluate the DSA module on both 2D and 3D backbone networks.
Specifically, the slow-only branch of SlowFast~\cite{slowfast} is applied as our 3D backbone baseline (denoted as I3D) and TSM~\cite{tsm} is applied as our 2D backbone baseline. 
We use random initialization for our DSA module and train the model in an end-to-end manner. 
Following the similar practice in V4D~\cite{zhang2020v4d} and TSN~\cite{tsn}, we uniformly divide the whole video into $U$ segments and we set $U=4$ in this paper, then randomly select a snippet of consecutive 32 or 64 frames from each segment as temporal augmentation. For each snippet, we uniformly sample 4 or 8 frames with a fixed stride of 8. The size of the short side of these frames is fixed to 256 and then random scaling is utilized for data augmentation. Finally, we resize the cropped regions to 224\x224 for network training. 
On the Mini-Kinetics-200 dataset, the learning rate is 0.01 and will be reduced by a factor of 10 at 40 and 60 epochs (70 epochs in total) respectively. 
On the Kinetics-400 dataset, our model is trained for 100 epochs starting with a learning rate of 0.01 and reducing it by a factor of 10 at 50, 75, and 90 epochs.
For Something-Something V1 dataset, our model is trained for 50 epochs starting with a learning rate 0.01 and reducing it by a factor of 10 at 30, 40 and 45 epochs.
For these three datasets, our models are initialized by pre-trained models on ImageNet~\cite{deng2009imagenet}. 
As for ActivityNet-v1.3, we followed the common practice to fine-tune from Kinetics-400 pre-trained weights and start training with a learning rate of 0.01 for 200 epochs. The learning rate is decayed by a factor of 10 at 80 and 160 epochs.


\textbf{Inference.} 
We follow the common setting in \cite{slowfast} to just uniformly sample multiple clips from a video along its temporal axis. We sample 8 clips unless otherwise specified. For each clip, we resize the short side to 256 pixels and take 3 spatial crops in each frame.

\subsection{Ablation Studies}
\label{sec:ab}
This section provides ablation studies on Mini-Kinetics-200 to investigate different aspects of the DSA module as shown in Table~\ref{tab:ablations}. Accordingly, concrete analysis is as follows.

\textbf{Effectiveness of DSA.} 
 We first investigate the performance of our DSA module with a toy backbone I3D ResNet-18 (R18). Specifically, we follow the Slow path in SlowFast~\cite{slowfast} to build the I3D R18 backbone which is the same as the structure in V4D~\cite{zhang2020v4d}. As it can be seen in Table~\ref{tab:ablation:r18_perf}, long-term modeling is complementary to clip-based method I3D R18. Specifically, when equipped with TSN, V4D, and our DSA, I3D R18 obtains 0.8\%, 3.4\%, and 5.1\% gain respectively. In comparison to these video-level methods, DSA significantly outperforms TSN and V4D by 4.3\% and 1.7\% accuracy with the same training protocols and even using less number of snippets for inference. In the following, unless stated otherwise, we apply I3D ResNet-50 (R50) as the backbone. 

\textbf{Inserted Position.}
As in Figure~2(b), there are four optional positions to integrate our DSA module into the standard I3D ResNet block. Table~\ref{tab:ablation:position} shows the results of DSA module in different position. We find that position \RNum{2} gets slightly higher than position \RNum{1} and \RNum{4}. Hence, the DSA module is inserted before the second convolution of the I3D ResNet Block by default in the following ablations.

\begin{table*}[th]
\caption{Comparison with the state-of-the-art models on Mini-Kinetics-200 dataset. $T$ denoted the temporal length of video snippet, $U$ denoted the number of video snippets.}
\begin{center}
\begin{tabular}{lccccc}
\shline
{ Method} & { Backbone} & { $T_{train} \times U_{train}$} & {$T_{infer} \times U_{infer} \times$ \#crop} & {Top-1} & {Top-5}\\
\hline
S3D~\cite{s3d} & {S3D Inception} & {$64 \times 1$} & {N/A}& 78.9\% & -\\
I3D~\cite{CGNLNetwork2018} &{3D ResNet50} & {$32 \times 1$} &{$32 \times 10 \times 3$} &75.5\% & 92.2\% \\
I3D~\cite{CGNLNetwork2018} &{3D ResNet101} & {$32 \times 1$} &{$32 \times 10 \times 3$} &77.4\% & 93.2\% \\
I3D+NL~\cite{CGNLNetwork2018} &{3D ResNet50} & {$32 \times 1$} &{$32 \times 10 \times 3$} &77.5\% & 94.0\% \\
I3D+CGNL~\cite{CGNLNetwork2018} & {3D ResNet50} & {$32 \times 1$} &{$32 \times 10 \times 3$} & 78.8\% & 94.4\% \\
I3D+NL~\cite{CGNLNetwork2018} & {3D ResNet101} & {$32 \times 1$} &{$32 \times 10 \times 3$} & 79.2\% & 93.2\% \\
I3D+CGNL~\cite{CGNLNetwork2018} & {3D ResNet101} & {$32 \times 1$} &{$32 \times 10 \times 3$} & 79.9\% & 93.4\% \\ \hline
V4D+I3D ~\cite{zhang2020v4d} &{3D ResNet18} & {$4 \times 4$} & {$4 \times 10 \times 3$} & 75.6\% &  92.7\% \\
V4D+I3D ~\cite{zhang2020v4d} &{3D ResNet50} & {$4 \times 4$} & {$4 \times 10 \times 3$} & 80.7\% &  95.3\% \\
\hline
DSA+I3D~(Ours) &{3D ResNet18} & {$4 \times 4$} & {$4 \times 8 \times 3$} & 77.3\% &  93.9\% \\
DSA+I3D~(Ours) &{3D ResNet50} & {$4 \times 4$} & {$4 \times 8 \times 3$} & \textbf{81.8\%} &  \textbf{95.4\%} \\
\shline
\end{tabular}
\end{center}
\label{cmp_minik200}
\end{table*}

\textbf{Parameter choices of $\bm{\alpha}$.}
We experiment with different $\alpha$ to figure out the optimal super-parameters of fully connected layers in the DSA module as shown in Figure 2(c). And in Table~\ref{tab:ablation:alpha}, our module with $\alpha=2$ get better performance thus we choose it in our experiments.

\begin{table*}[th]
\centering
\caption{Comparison with the state-of-the-art models on Kinetics-400 dataset. We report the inference cost by computing the GFLOPs. $^*$ indicates the result of our calculation using the official model.}
\scalebox{1}{
\begin{tabular}{lcccccc}
\shline
Method &  Backbone &  {$T_{infer} \times U_{infer} \times$ \#crop} &  GFLOPs &  Top-1 &  Top-5\\
\hline
 
TSM~\cite{tsm} & ResNet-50 & 8\x10\x3 & 33\x30$=$990 & 74.1\% & 91.2\%\\
TEINet~\cite{teinet} & ResNet-50 & 8\x10\x3 & 33\x30$=$990 & 74.9\% & 91.8\% \\ 
TEA~\cite{li2020tea} & ResNet-50& 8\x10\x3 & 35\x30$=$1050 & 75.0\% & 91.8\% \\
TANet~\cite{liu2020tam} & ResNet-50 & 8\x10\x3 & 43\x30$=$1290 & 76.1\% & 92.3\% \\
MVFNet~\cite{wu2020MVFNet} & ResNet-50& 8\x10\x3 & 33\x30$=$990 & 76.0\% & 92.4\% \\
NL+I3D~\cite{nonlocal} & 3D ResNet-50 & 32\x10\x3 & 70.5\x30$=$2115 & 74.9\% & 91.6\% \\
NL+I3D~\cite{nonlocal} & 3D ResNet-50 & 128\x10\x3 & 282\x30$=$8460 & 76.5\% & 92.6\% \\ 
X3D-L~\cite{feichtenhofer2020x3d} & - & 16\x10\x3 & 24.8\x30$=$744  & 77.5\% & 92.9\% \\
Slowfast~\cite{slowfast} & 3D R50+3D R50 & (4+32)\x10\x3 & 36.1\x30$=$1083 & 75.6\% & 92.1\% \\
Slowfast~\cite{slowfast} & 3D R50+3D R50 & (8+32)\x3\x10 & 65.7\x30$=$1971 & 77.0\% & 92.6\% \\
Slowfast~\cite{slowfast} & 3D R101+3D R101 & (8+32)\x3\x10 & 106\x30$=$3180 & 77.9\% & 93.2\% \\
Slowonly~\cite{slowfast} & 3D ResNet-50 & 8\x10\x3 & 41.9\x30$=$1257 & 74.9\% & 91.5\% \\ \hline
V4D+I3D~\cite{zhang2020v4d}  & 3D ResNet-50 & 8\x10\x3 & 286.1\x2.5\x3$=$2146$^*$ & 77.4\% & 93.1\% \\ \hline



DSA+I3D (Ours) & 3D ResNet-50& 4\x8\x3 & 83.8\x2\x3$=$503 & \textbf{77.7\%} & \textbf{93.1\%} \\ 
DSA+I3D (Ours) & 3D ResNet-50& 8\x8\x3 & 167.7\x2\x3$=$1006 & \textbf{78.2\%} & \textbf{93.2\%} \\
DSA+I3D (Ours) & 3D ResNet-50& (4+8)\x8\x3 & 251.5\x2\x3$=$1509 & \textbf{79.0\%} & \textbf{93.7\%} \\


\shline
\end{tabular}
}
\label{tab:sota_kinetics}
\end{table*}

\textbf{Which stage to insert DSA blocks.} 
We denote the conv2\_x to conv5\_x of ResNet architecture as res\{2\} to res\{5\}.
To evaluate which stage may be important for DSA design, we insert DSA blocks into a different stage of I3D R50 and only replace one with DSA block for every two I3D ResNet blocks. Table~\ref{tab:ablation:stage} shows that the performance of DSA blocks on res\{2\}, res\{3\}, res\{4\} or res\{5\} is similar, which suggests that DSA can perform effective adaptive fusion on both low-level features and high-level features of snippets.

\textbf{Parameter choices of $\bm{\beta}$.} 
As shown in Table~\ref{tab:ablation:beta}, we compare networks with different proportion of input channels ($\beta=1/8, 1/4, 1/2, 1$) for long-term temporal modeling. Here we add DSA blocks into res\{5\} and we observe that the change in $\beta$ appeared to have little impact on performance thus we adopt $\beta=1/8$ for efficiency in the following. 

\textbf{The number of DSA Blocks.} 
We here expect to figure out if more DSA blocks can further improve performance. Table~\ref{tab:ablation:multi-stage} compares DSA blocks added to multiple stages of I3D ResNet. The res\{3,4\} achieves the highest performance and will be used in the following experiments.

\textbf{Generality of DSA.} 
Being a general module, DSA could be combined with both 2D and 3D clip-based networks. 
To show this, we add DSA to TSM (DSA+TSM R50) which is the recent state-of-the-art model based on 2D CNN and train such a combination in an end-to-end manner. 
For the 3D backbone, the slow-only branch of SlowFast is applied as our backbone network (denoted as I3D R50) due to its promising performance on various datasets.
As shown in Table~\ref{tab:ablation:short-term}, adding DSA to TSM R50 could boost the top-1 accuracy by 3.0\%. A consistent improvement is observed as in I3D R50, which has an increase of 3.8\%.

\textbf{Different Backbone.}
In Table~\ref{tab:ablation:backbone} we compare various instantiations of DSA networks on Mini-Kinetics-200. 
As expected, using deeper backbones is complementary to our method. Comparing with the I3D R18 counterparts, our DSA gets additional improvement on I3D R50. 

\textbf{Comparison with SENet.}
To avoid misunderstanding, here we clarify the essential difference between SENet and DSANet is that SENet is a kind of \textbf{self-attention} mechanism by using its global feature to calibrate the features in channel-level, while our DSA generates \textbf{dynamic convolution kernel} to adaptively aggregate the snippet-level features with a global view.
To prove the effectiveness of DSA, we conduct the comparative experiments in Table~\ref{tab:ablation:backbone}. Under the same setting, our DSA outperforms the SE module with a clear margin (\textbf{77.3\%} \emph{vs.} 73.8\%, \textbf{81.8\%} \emph{vs.} 78.5\%) on I3D R18 and I3D R50.

\textbf{Training Cost.}
Here we make a comparison with the state-of-the-art video-level method V4D+I3D under the same setting of backbone and inputs. 
We list the number of FLOPs for all models in Table~\ref{tab:ablation:flops}, our DSA+I3D is more lightweight (83.8G \emph{vs.} 143.0G) than V4D+I3D~\cite{zhang2020v4d} and achieve a similar cost with the baseline TSN+I3D.

\subsection{Comparison with State-of-the-arts}
In this section, we compare our approach with the state-of-the-art methods on trimmed video datasets (\emph{i.e.,} Mini-Kinetics-200 \cite{s3d}, Kinetics-400 \cite{kay2017kinetics} and Something-Something V1 \cite{sth-sth}) and untrimmed video dataset (\emph{i.e.,} ActivityNet~\cite{caba2015activitynet}).

\textbf{Results on Mini-Kinetics-200.} 
We evaluate the DSA-equipped networks on Mini-Kinetics-200. 
As shown in Table~\ref{cmp_minik200}, DSA with less input already outperforms the previous methods. Notably, DSA-R18 (77.3\%) has similar accuracy with I3D R101 (77.4\%) and I3D R50-NL (77.5\%). Compared with V4D, DSA achieves better accuracy on R18 (\textbf{77.3\%} \emph{vs.} 75.6\%) and R50 (\textbf{81.8\%} \emph{vs.} 80.7\%) with fewer clips for prediction (\textbf{8} \emph{vs.} 10) which shows the superiority of our method.

\begin{figure}[t]
    \centering
    \includegraphics[width=0.48\textwidth]{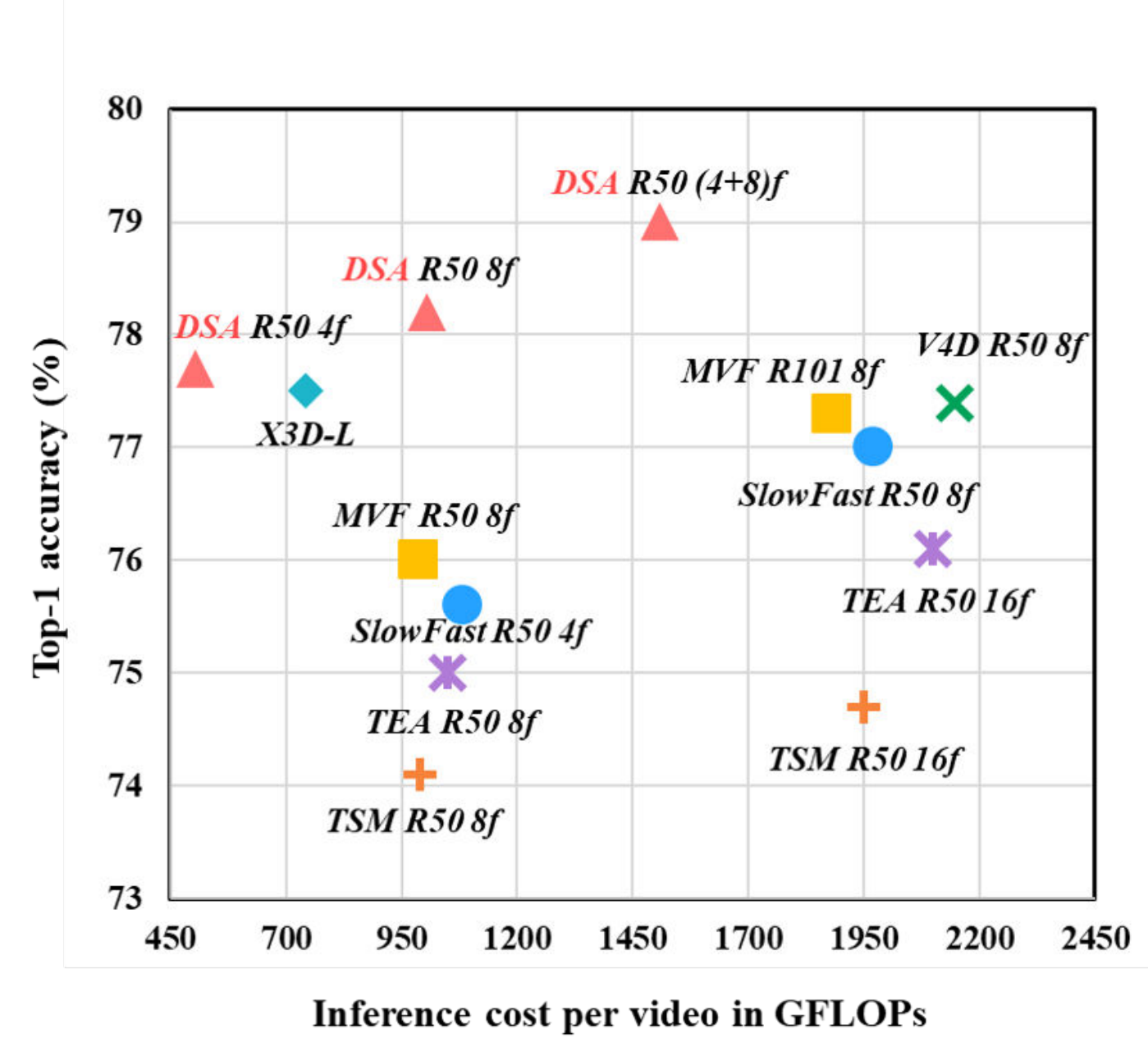}
    \caption{Accuracy-computation trade-off on Kinetics-400 for different methods in the inference phase. The horizontal axis shows the inference cost per video.}   
    \vspace{-8pt}
    \label{fig:cmp_sota}
\end{figure}

\textbf{Results on Kinetics-400.}
We make a thorough comparison in Table~\ref{tab:sota_kinetics}, where our DSA outperforms the recent SOTA approaches on Kinetics-400. Here we only list the models using RGB as inputs for fair comparisons. We provide results of our DSA network trained with two different inputs (\emph{i.e.}, 4 frames ($T=4$) or 8 frames ($T=8$) sampled from each snippet).
The upper part of the Table~\ref{tab:sota_kinetics} shows the results of clip-based models. Compared with the 2D CNN-based model, during inference, when utilizing less frames as input (4\x8 \emph{vs.} 8\x10), our DSA still outperforms these current state-of-the-art efficient methods (\emph{e.g.}, TSM, TEINet, TEA and MVFNet) with a clear margin (\textbf{77.7}\% \emph{vs.} 74.1\%/74.9\%/75.0\%/76.0\%) meanwhile with (2\x) less calculation cost. 
Compared with computationally expensive models of 3D architecture, DSA with 4$\times$8 frames input outperforms NL I3D-R50 with 32$\times$10 frames (\textbf{77.7}\% \emph{vs.} 74.9\%) but only uses \textbf{4.2\x} less GFLOPs, and uses \textbf{16.8\x} less GFLOPs than NL I3D-R50 with 128$\times$10 frames meanwhile achieves a better accuracy (\textbf{77.7}\% \emph{vs.} 76.5\%). 
Moreover, DSA with 8$\times$8 frames input achieves a better accuracy than Slowfast-R50 and Slowfast-R101 (\textbf{78.2}\% \emph{vs.} 77.9\%/77.0\%) when using \textbf{3.2\x}, \textbf{2\x} less GFLOPs respectively.
Compared with X3D~\cite{feichtenhofer2020x3d}, our DSA achieves better performance (\textbf{77.7\%} \emph{vs.} 77.5\%) while using less inference cost (\textbf{503G} \emph{vs.} 744G).
Remarkably, equipped with the DSA module, the top-1 accuracy of Slowonly (namely I3D R50 above) improves from 74.9\% to \textbf{78.2\%} on Kinetics-400.
Then compared with video-level method, our DSA with 4\x8 frames as input outperform V4D with 8\x10 frames (\textbf{78.2}\% \emph{vs.} 77.4\%) when using much less GFLOPs (\textbf{4.3\x}). 
Furthermore, our DSA has better results with longer inputs, and exceed V4D by 0.8\%. 
For readers’ reference, here we also report the results of ensemble the two models.
As shown in Figure~\ref{fig:cmp_sota}, comparison with prior works (SlowFast~\cite{slowfast}, X3D~\cite{feichtenhofer2020x3d}, TSM~\cite{tsm}, TEA~\cite{li2020tea}, MVFNet~\cite{wu2020MVFNet}, V4D~\cite{zhang2020v4d}), our DSANet achieves state-of-the-art performance on Kinetics-400 and get better accuracy-computation trade-off than other methods.


\begin{table}[th]
\centering
\caption{Comparison with state-of-the-art on ActivityNet v1.3. The results using Multi-scale Temporal Window Integration~\cite{tsn} (abbreviated as M-TWI) with only RGB frames.}

\begin{tabular}{l|l|l}
\shline
{Model}& Backbone & mAP\\
\hline
TSN~\cite{tsn} & BN-Inception & 79.7\% \\
TSN~\cite{tsn}& Inception V3 & 83.3\% \\
TSN-Top3~\cite{tsn}& Inception V3 & 84.5\% \\
V4D+I3D~\cite{zhang2020v4d} & {3D ResNet50} & 88.9\%  \\
\hline
DSA+I3D (Ours) & {3D ResNet50} & \textbf{90.5\%} \\
\shline


\end{tabular}
\label{cmp anet}
\end{table}

\textbf{Results on ActivityNet-v1.3.}
To verify the generalization ability of our DSANet, we further evaluate the performance of our method on a large-scale untrimmed video recognition benchmark, ActivityNet~\cite{caba2015activitynet}. 
We finetuned the model pre-trained on Kinetics-400 on the Activitynet-v1.3 dataset and report the mean average precision (mAP) follow the official evaluation metrics. 
As shown in Table~\ref{cmp anet}, our method outperforms the previous methods (\emph{i.e.,} V4D, TSN-Top3, TSN) with a clear margin (90.5\% vs. 88.9\%, 84.5\%, 83.3\%).

\begin{table}[h]
\caption{Comparison with the state-of-the-art models on Something-Something V1 dataset.}
\begin{center}
\begin{tabular}{lcc}
\shline
{ Method} & { Backbone} &  {Top-1} \\
\hline
MultiScale TRN~\cite{trn} & BN-Inception & 34.4\% \\
ECO~\cite{eco} & BN-Inception+3D ResNet 18 & 46.4\% \\
S3D-G~\cite{s3d} & {S3D Inception} & 45.8\% \\
Nonlocal+GCN~\cite{gcn} &{3D ResNet50} & 46.1\%  \\
TSM~\cite{tsm} & {ResNet50} & 47.2\% \\
I3D (our impl.) & {3D ResNet50} & 48.7\% \\
V4D+I3D ~\cite{zhang2020v4d} &{3D ResNet50}  & 50.4\%  \\
\hline
DSA+I3D~(Ours) &{3D ResNet50}  & \textbf{51.8\%}  \\
\shline
\end{tabular}
\end{center}
\label{cmp_sthv1}
\end{table}

\begin{figure*}[t]
    \centering
    \includegraphics[width=0.95\textwidth]{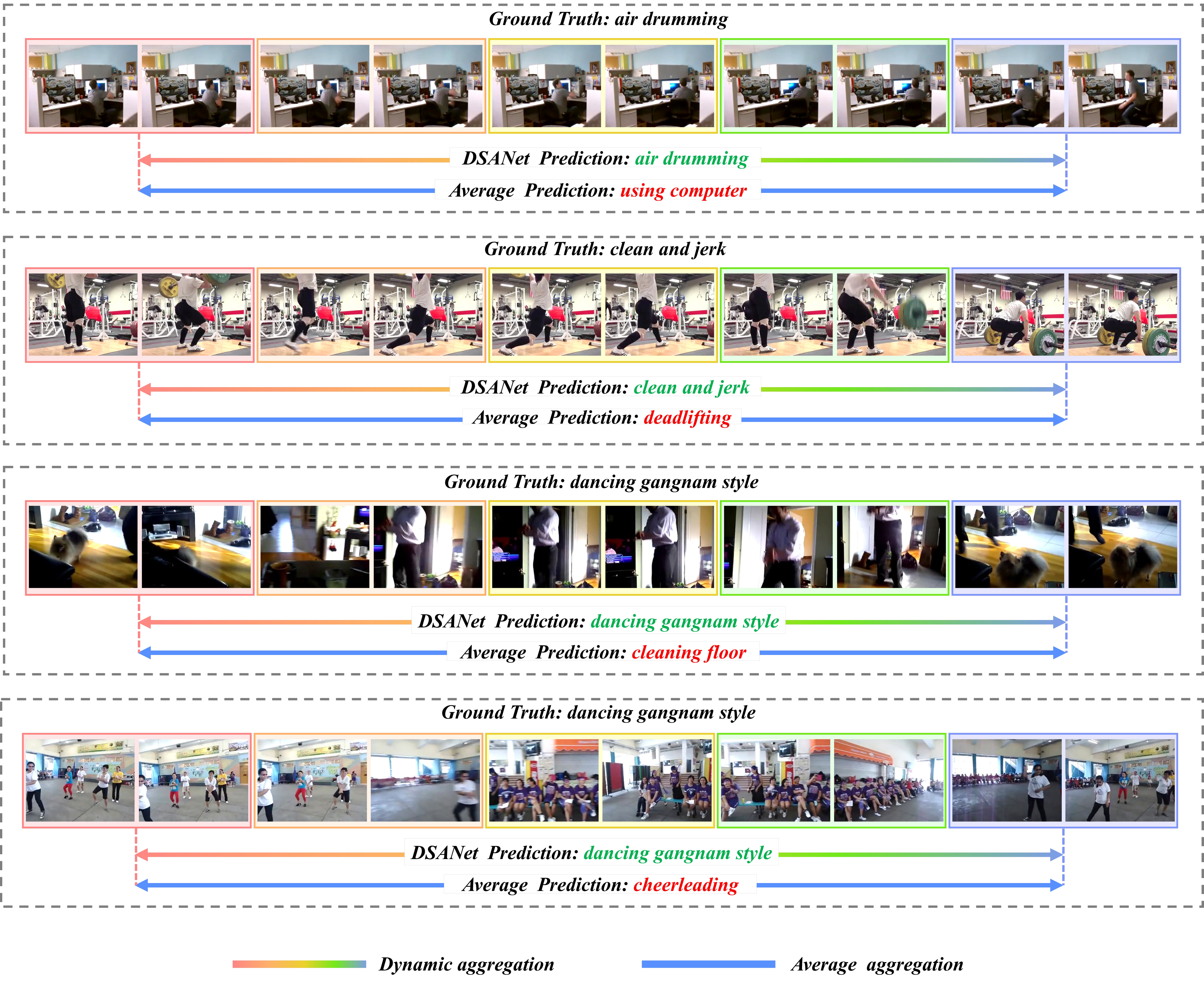}
    \caption{Visualization of predictions with DSANet and I3D.  (\textcolor[RGB]{0,176,80}{Green}: correct predictions,  \textcolor[RGB]{246,0,0}{Red}: wrong predictions.)}
    \label{fig:vis}
\end{figure*}

\textbf{Results on Something-Something V1.}
We also evaluate our method on Something-Something V1 dataset which focuses on modeling temporal relation and is more complicated than Kinetics dataset.
As shown in Table~\ref{cmp_sthv1}, comparing with prior works, our DSA obtains comparable or better performance.
To verify the effectiveness of DSA, we report the baseline result of I3D ResNet-50 in our implementation for fair comparison. 
Remarkably, equipped with the DSA module, the top-1 accuracy of I3D ResNet-50 improves from 48.7\% to \textbf{51.8\%} on Something-Something V1.
In this way, we believe our DSA can achieve higher accuracy when using better snippet-level counterpart.

\section{Visualization}
We also visualize some examples of predictions generated by DSANet and I3D model on the validation data of Kinetics-400 in Figure~\ref{fig:vis}.
Compared with I3D which simply averaging the prediction scores of all snippets to yield a video-level prediction, we see that our DSANet can learn global context information and adaptively capture relation-ship among snippets via dynamic aggregation, then provide more accurate results with contextual perspective.

\section{Conclusion}
In this paper, we presented the Dynamic Segment Aggregation (DSA) module to better exploit long-range spatiotemporal representation in videos.
The DSA module can be easily applied to the existing snippet-level methods in a plug-and-play manner with a negligible computational cost, bringing consistent improvements on both 2D/3D CNN backbone.
A series of empirical studies verify that our DSA network is an accurate and efficient model for large scale video classification.
The experimental results show that our method has achieved state-of-the-art performance on four public benchmarks with a high inference efficiency.

\bibliographystyle{ACM-Reference-Format}
\balance
\bibliography{sample-base}

\end{document}